\documentclass{article} 
\usepackage{iclr2024_conference,times}

\usepackage{amsmath,amsfonts,bm}

\def\eqref#1{equation~\ref{#1}}

\def\1{\bm{1}}

\def\ru{{\textnormal{u}}}

\def\vh{{\bm{h}}}

\def\mW{{\bm{W}}}

\def\mPhi{{\bm{\Phi}}}

\DeclareMathAlphabet{\mathsfit}{\encodingdefault}{\sfdefault}{m}{sl}
\SetMathAlphabet{\mathsfit}{bold}{\encodingdefault}{\sfdefault}{bx}{n}

\newcommand{\R}{\mathbb{R}}

\usepackage{hyperref}
\usepackage{url}

\usepackage{amsmath,amssymb,amsfonts}
\usepackage{graphicx}
\usepackage{algorithm}
\usepackage{algorithmic}
\usepackage{booktabs,multirow,makecell}

\def\mr[#1]#2#3{\multirowcell{#2}[#1]{#3}}
\usepackage{booktabs,multirow,makecell}
\def\BibTeX{{\rm B\kern-.05em{\sc i\kern-.025em b}\kern-.08em
    T\kern-.1667em\lower.7ex\hbox{E}\kern-.125emX}}

\title{Customizable Combination of Parameter-Efficient Modules for Multi-Task Learning}

\author{Haowen Wang \\
AntGroup\\
Shanghai, China \\
\texttt{wanghaowen.whw@antgroup.com} \\
\And
Tao Sun \\
AntGroup\\
Shanghai, China \\
\texttt{suntao.sun@antgroup.com} \\
\AND
Cong Fan \\
AntGroup\\
Shanghai, China \\
\texttt{fancong.fan@antgroup.com}
\AND
Jinjie Gu \\
AntGroup\\
Shanghai, China \\
\texttt{jinjie.gujj@antgroup.com}
}

\iclrfinalcopy 
\begin{document}

\maketitle

\begin{abstract}
Modular and composable transfer learning is an emerging direction in the field of Parameter Efficient Fine-Tuning, as it enables neural networks to better organize various aspects of knowledge, leading to improved cross-task generalization.
In this paper, we introduce a novel approach Customized Polytropon ($\texttt{C-Poly}$) that combines task-common skills and task-specific skills, while the skill parameters being highly parameterized using low-rank techniques.
Each task is associated with a customizable number of exclusive specialized skills and also benefits from skills shared with peer tasks.
A skill assignment matrix is jointly learned.
To evaluate our approach, we conducted extensive experiments on the Super-NaturalInstructions and the SuperGLUE benchmarks.
Our findings demonstrate that $\texttt{C-Poly}$ outperforms fully-shared, task-specific, and skill-indistinguishable baselines, significantly enhancing the sample efficiency in multi-task learning scenarios.
\end{abstract} 

\section{Introduction}
\label{introduction}

As the number of parameters in Large Language Models (LLMs) continues to grow, training these models efficiently with limited computational resources has become a challenge.
In recent years, there has been a shift towards employing Parameter Effective Fine-Tuning (PEFT) methods to address this issue.
Examples of such methods include LoRA~\citep{hu2022lora}, AdaLoRA~\citep{zhang2023adaptive}, and (IA)$^3$~\citep{liu2022few}. These methods focus on fine-tuning the adapter while freezing the pre-trained model, effectively reducing the computational cost.
By selectively updating only a portion of the model parameters, PEFT methods enable efficient training and utilization of large foundation models.
This line of approaches allows for more effective use of resources while maintaining the performance of the pre-trained model on downstream tasks~\citep{hu2022lora}. 
However, despite the popularity and widely adoption of PEFT methods, the learning effectiveness of such methods, especially in multi-task scenarios, is under explored.

LLMs are famous for their extraordinary capabilities on solving multiple tasks in zero-shot or few-shot manners~\citep{DBLP:journals/corr/abs-2005-14165}. Basic PEFT methods mentioned earlier don't take the multitask essence of real-world data into account and rely heavily on the base foundation model's capacities on the multitask generalization.
Building upon the basic PEFT methods, various training approaches designed for Multi-Task Learning (MTL) have been proposed~\citep{pfeiffer2020adapterfusion,vu2021spot, asai2022attempt,chronopoulou2023adaptersoup,zadouri2023pushing}.
One simple solution is to perform multitask training by training a large model on a combination of multiple tasks. 
This involves training the model on the union of training tasks and subsequently evaluating its performance on different testing tasks ~\citep{ye2021crossfit, liu2022few}. However, this approach overlooks the relationships between the tasks and is vulnerable to negative transfer, where the gradients associated with different tasks are misaligned ~\citep{wang2020gradient}. This misalignment of gradients can lead to sub-optimal performance and hinder the effective utilization of learned knowledge across tasks.

To enhance sample efficiency, MoLoRA~\citep{zadouri2023pushing} has been introduced. MoLoRA successfully applies the Mixture-of-Expert (MoE) architecture to PEFT methods and improves the model's generalization capacity across various tasks by jointly learning multiple LoRA instances. MoLoRA views each LoRA as a lightweight expert, following MoE framework, and thus allows for more specialized adaptation to different tasks from shared knowledge learnt through parallel instances.

In a recent work, ~\citet{ponti2022combining} developed Polytropon ($\texttt{Poly}$) to tackle the challenges associated with multitask learning.
The central idea of $\texttt{Poly}$ is to consider each task-specific adapter as a composition of reusable skillset of basic adapters or modules. Specifically, $\texttt{Poly}$ jointly learns an inventory of adapters (for example, LoRA) and a simple routing vector that selects and combines a (variable-size) subset of adapters for each individual task.
This approach significantly improves the efficiency of sample sharing and utilization between multiple tasks.
Although not mentioned explicitly in the original work~\citep{ponti2022combining}, we argue that each adapter in $\texttt{Poly}$ can be viewed as a lightweight expert and then the whole structure follows the same pattern as Multi-gate Mixture-of-Experts (MMoE)~\citep{10.1145/3219819.3220007}, a well-known MTL framework. To be noted, in \citet{ponti2022combining} also introduced a structure MoE-LoRA when treating all tasks the same, which is a simplified version of MoLoRA where the routing is controlled by a function of hidden states instead.

Based on $\texttt{Poly}$, a subsequent research titled Multi-Head Routing ($\texttt{MHR}$)~\citep{caccia2022multi} was proposed. $\texttt{MHR}$ extends the concept of parallel thinking to low-rank decomposition and introduces further improvements to the basic unit of the adapter.
By leveraging low-rank decomposition, $\texttt{MHR}$ enhances the model's ability to generalize across different tasks. The parallel improvements in the adapter's basic units allow for more efficient adaptation to different tasks while still benefiting from shared knowledge.

\begin{figure}[htbp]
\begin{center}
\includegraphics[width=\textwidth]{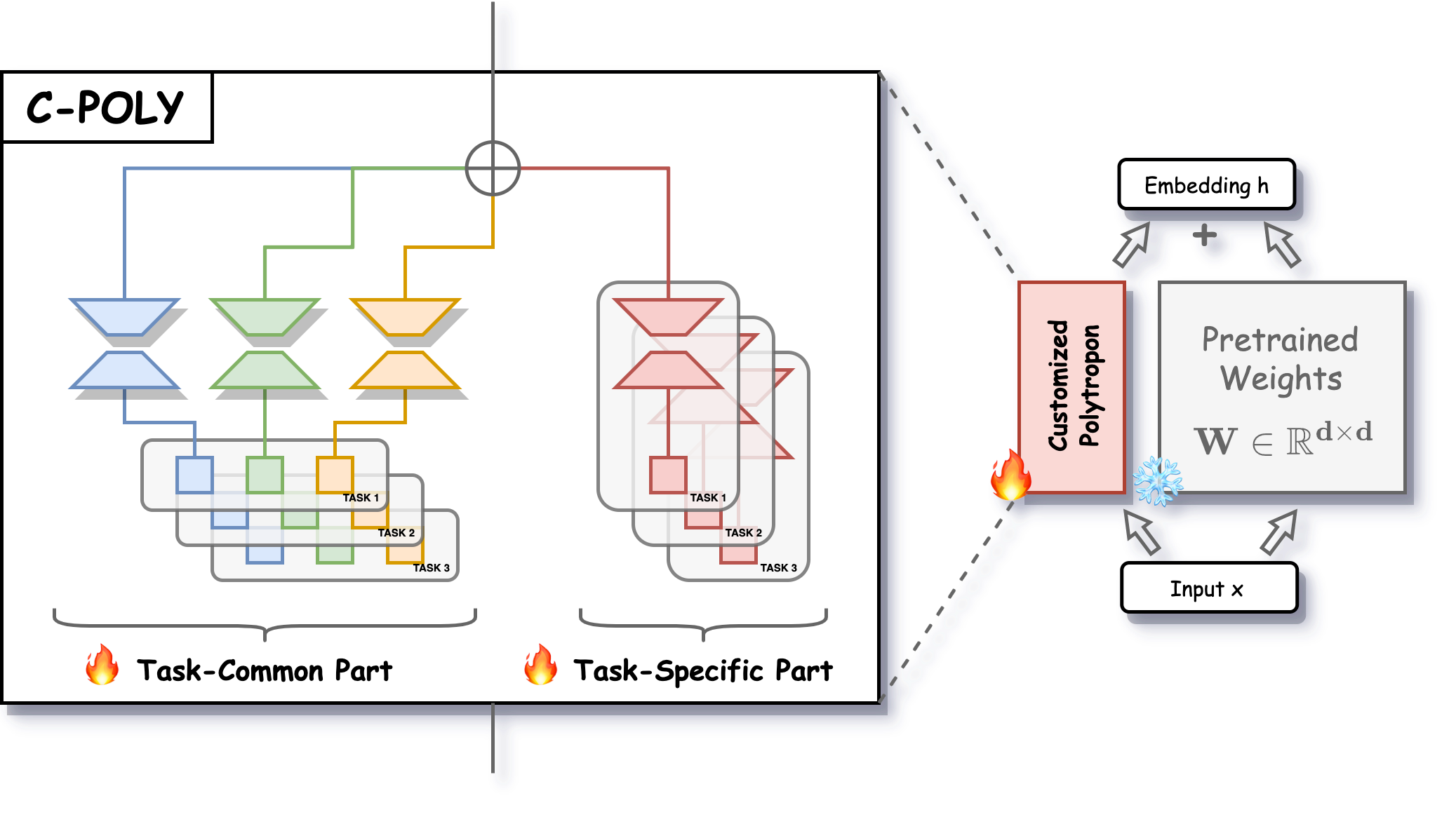}
\end{center}
\caption{Overview of Customized Polytropon ($\texttt{C-Poly}$) framework}
\label{fig1}
\end{figure}

Inspired by Customized Sharing Control (CGC) and Progressive Layered Extraction (PLE)~\citep{10.1145/3383313.3412236}, our research has made additional improvements over $\texttt{Poly}$.
We assume that in MTL, many tasks do share transferable knowledge, while each of them requires discriminative abilities.
Based on this assumption, we explicitly divide modular skills into task-common skill modules and task-specific skill modules.
We propose \textbf{Customized Polytropon ($\texttt{C-Poly}$)}, where for each task, two components are learnt jointly, a task-specific skill module and a task-common skill module.
This allows each task to be characterized by not only a shared task-common module as in $\texttt{Poly}$ and $\texttt{MHR}$, but also a unique subset of skills, mitigating the effect of negative transferring and leading to improved multi-task performance.
Furthermore, $\texttt{C-Poly}$ promotes interpretability by learning an explicit hierarchy of tasks based on the skills they select, which provides insights into the relationship between tasks and the skills needed to perform them effectively.
Additionally, we observe the task-specific instantiation of a neural network by combining specific parameters of the active skills.
This reduces memory usage and facilitates more effective training, as only the necessary parameters associated with active skills are utilized.
Overall, our approach enhances the efficiency and effectiveness of modular skill multi-task learning, enabling better performance, interpretability, and resource utilization.

In Section~\ref{sec:experiments}, we conducted extensive experiments on multiple datasets and model architectures. The results of these experiments have consistently demonstrated our method $\texttt{C-Poly}$ has surpassed the performance of the most advanced PEFT methods and achieved state-of-the-art (SOTA) results in different multitasking scenarios.

\section{Methodology}
\label{sec:methodology}

The proposed unified MTL framework $\texttt{C-Poly}$ is shown in Figure~\ref{fig1}, which aims to enhance sample efficiency for each task by leveraging strengths from all other tasks while keeping task-specific abilities.
Suppose there are $T$ tasks and for each task, task-specific data input $x^t$, $t\in\{1, 2, \dots,T\}$. The MoE-like structure consists of adapter modules (or experts), $\mPhi = \{ \phi_1, \phi_2, \dots, \phi_{|\mPhi|}\}$ and each adapter $\phi_i$ can be regarded as a function of the input data $x^t$.
The major improvement of $\texttt{C-Poly}$ is to explicitly categorize adapter modules $\mPhi$ into two separate parts:
\begin{itemize}
    \item A task-common skillset: $\mPhi_{A} = \{ \phi_1, \phi_2, \dots, \phi_{A}\}$ having $A$ adapters.
    \item A task-specific skillset: $\mPhi^{t}_{B} = \{ \phi^{t}_{1}, \phi^{t}_{2}, \dots, \phi^{t}_{B}\}$ having $B$ adapters for each task $t$.
\end{itemize}
In total, there are $|\mPhi_A| + T \times |\mPhi_B^t| = A + T \times B$ adapters. For a simplified yet generalizable discussion, we would set $B=1$ to keep only one task-specific adapter in the following experiments.

The combined output of the $\texttt{C-Poly}$ adapter modules for each task input $x^t$ can be expressed in Equation~\ref{eq:cpoly} with $w_i$ representing the learnable weight of each adapter's output.

\begin{equation}
\label{eq:cpoly}
        \underbrace{\sum_{i=1}^{A}{w^{t}_{i}\phi_{i}(x^t)}}_{\text{Task-Common}} + \underbrace{\sum_{j=1}^{B}w^{t}_{j}\phi^{t}_{j}(x^t)}_{\text{Task-Specific}}
        = \sum_{i=1}^{A}{w^{t}_{i}\phi_{i}(x^t)} + 
        w^{t}\phi^{t}(x^t)
\end{equation}

In the task-common part, the set of adapters $\phi_{i}$ are shared across all tasks, while the weights $w^{t}_{i}$ are exclusive to each task $t$. In contrast, both the weights $w^{t}_{i}$ and adapters $\phi^{t}_{i}$ are customized for each individual task $t$ in the task-specific part.

Following the notation above, various MoE-like PEFT structures can be mathematically formulated together in Table~\ref{moe_peft}. Both the MoE and MMoE models only consist of the task-common part of Equation~\ref{eq:cpoly}: in the conventional MoE approach, tasks are not differentiated, leading to shared parameters across all tasks; the MMoE framework assigns task-specific weights for each individual task, while still maintaining a shared pool of experts or adapters.

\begin{table}[htbp]
\caption{Comparison between different MoE-like PEFT methods}
\label{moe_peft}
\begin{center}
\begin{tabular}{ccc}
\toprule
\multicolumn{1}{c}{\bf MoE Structures}  &\multicolumn{1}{c}{\bf PEFT Methods}    &\multicolumn{1}{c}{\bf Task Output}
\\ \midrule
Conventional MoE
&\makecell[c]{
MoLoRA~\citep{zadouri2023pushing},\\
MoE-LoRA~\citep{ponti2022combining}
}
&$\displaystyle\sum_{i=1}^{A}{w_{i}\phi_{i}(x^t)}$
\\ \midrule
MMoE~\citep{10.1145/3219819.3220007}
&\makecell[c]{
$\texttt{Poly}$~\citep{ponti2022combining},\\
$\texttt{MHR}$~\citep{caccia2022multi}
}
&$\displaystyle\sum_{i=1}^{A}{w^{t}_{i}\phi_{i}(x^t)}$
\\ \midrule
\makecell[c]{
CGC~\citep{10.1145/3383313.3412236}, \\
PLE~\citep{10.1145/3383313.3412236}
}
&\textbf{Our Method} $\texttt{C-Poly}$
&$\displaystyle\sum_{i=1}^{A}{w^{t}_{i}\phi_{i}(x^t)} + w^{t}\phi^{t}(x^t)$\\ \bottomrule
\end{tabular}
\end{center}
\end{table}

In $\texttt{C-Poly}$, the weights associated with each adapter can be represented together as one allocation matrix $\mW \in \R^{T \times (A + T)}$ when $B=1$. This matrix can be further decomposed into two distinct components: $\mW_A \in \R^{T \times A}$ and $\mW_B\in \R^{T \times T}$:
\begin{align}
    \mW 
    &= \left[
    \begin{array}{c|c}
        \mW_A & \mW_B
    \end{array}
    \right] \\
    & = \left[
    \begin{array}{cccc|cccc}
	w_{1}^{1}&w_{2}^{1}&\cdots&w_{A}^{1}&w^{1}&0&0&0\\
	w_{1}^{2}&w_{2}^{2}&\cdots&w_{A}^{2}&0&w^{2}&\cdots&0\\
	\vdots&&\ddots&\vdots&\vdots&&\ddots&\vdots\\
	w_{1}^{T}&w_{2}^{T}&\cdots&w_{A}^{T}&0&0&\cdots&w^{T}\\
	\end{array}
    \right]
\end{align}

To optimize the learning process in skill acquisition, we have employed different learning methods for each component of the allocation matrix. Additionally, we have incorporated low-rank approximations to further enhance the parameter efficiency.

\subsection{Task-Common Skills Learning}
\label{subsec:task-common-skills-learning}
Task-common skills are universally applicable skills that all tasks can leverage. Previous research has focused on identifying the effectiveness of general skills modules for specific tasks by employing modular concepts at the structured input level inspired by cognitive mechanisms~\citep{bengio2017consciousness, ponti2022combining, caccia2022multi, zadouri2023pushing}. This concept has been translated into Softmax for cross-module and top-k selection in practical implementation.

Therefore, following~\citet{ponti2022combining} and \citet{caccia2022multi}, we utilize a task-common allocation matrix $\mW_{A} \in \{0, 1\}^{T \times A}$ with a uniform initialization. This matrix is employed to achieve the soft partitioning of general skills. Each element $w_{i}^{t}$ in $\mW_A$ is a binary value that indicates whether a particular task $t$ activates the adapter module $\phi_{i}$. However, since discrete binary matrices like $\mW_A$ are non-differentiable, learning cannot be accomplished through gradient descent. To overcome this limitation, we adopt the Gumbel-sigmoid approach~\citep{maddison2016concrete,jang2016categorical}, which allows us to obtain a set of continuously relaxed Bernoulli distributions. This approach guarantees both randomness and the ability to perform differentiable sampling:

\begin{equation}
    \hat{w_{i}^{t}}=\sigma \left[\log\frac{\sigma(w_{i}^{t})\ru}{(1-\sigma(w_{i}^{t}))(1-\ru)} \right],\quad
    \ru \sim \mathcal{U}(0, 1)
\end{equation}

\subsection{Task-Specific Skills Learning}
\label{subsec:Specialized Skills Learning}
Specialized skills refer to modular skills that acquire the distinctive attributes of each task. In complex and interconnected multitasking scenarios, a seesaw phenomenon commonly arises. In multitasking learning mode, there is often a trade-off between enhancing specific tasks' effectiveness and compromising others' effectiveness~\citep{wang2020gradient}. To address this, we explicitly differentiate between shared and exclusive skill modules. This separation allows us to amplify the inherent characteristics of individual tasks.

We initialize the task-specific skill allocation $\mW_{B} \in \R^{T \times T}$ as a unit diagonal matrix. Notably, during the actual training process, although only the diagonal of $\mW_{B}$ are weights we concerned, entries off the diagonal are also subject to potential updates. These side-effects indicate that while solving the current task, the exclusive specialized skills of other tasks can be leveraged without influencing the parameter values of specialized skills for those tasks.



\subsection{Parameter Efficiency}
We accomplished an efficient parameterization of skill modules by employing low-rank techniques. Every adapter in our $\texttt{C-Poly}$ experiments is Low-Rank Adapter (LoRA)~\citep{hu2022lora}. LoRA is a straightforward yet effective structure specifically tailored for Transformer-based models~\citep{10.5555/3295222.3295349}. The idea behind LoRA is relatively straightforward. 
LoRA decomposes each weight matrix of the linear transformation in Transformers into the multiplication of two low-rank matrices.
In other words, the linear projection $f: \R^{d} \rightarrow \R^{d}$ can be represented as follows, disregarding the bias part:

\begin{equation}
\vh_{l+1}=\vh_{l}\left[ \mW_l+ \Delta\mW \right] =\vh_{l}\left[ \mW_l+\mW_{down}\mW_{up} \right]
\end{equation}

Instead of $\Delta\mW \in \R^{d \times d}$, two much smaller matrices $\mW_{down} \in \R^{d \times r}$ and $\mW_{up} \in \R^{r \times d}$, with $r \ll d$, are obtained through gradient descend optimization.
Through the adoption of LoRA, updating each linear layer in the model only requires $2 \times r \times d$ parameters in the calculation, as opposed to the original $d \times d$ parameters. This results in a notable enhancement in parameter efficiency, enabling faster training even with limited computing resources.

As examined in~\citet{hu2022lora}, LoRA can be applied to various components of Transformers, such as query, key, value, and feed-forward layers, while the choice of rank $r$ does not hold significant importance. This suggests that LoRA exhibits versatility in its applicability. In our experiments, we replaced all query, key, and value layers as $\texttt{C-Poly}$, a combination of multiple LoRAs.


\section{Empirical Experiments}
\label{sec:experiments}

\subsection{Experimental Setup}
In order to evaluate the efficacy of our proposed unified MTL framework $\texttt{C-Poly}$, which incorporates both task-common and task-specific skills, we conducted experiments on two publicly available multitasking benchmarks: SuperGLUE~\citep{wang2019superglue} and Super Natural-Instructions (Super NI)~\citep{wang-etal-2022-super}.
The SuperGLUE benchmark is a widely acceptable benchmark for evaluating general-purpose language understanding. In our experiments, seven distinct tasks were selected from the benchmark that can be effectively evaluated using the accuracy metric.
The Super NI dataset, as a meta-dataset~\citep{DBLP:journals/corr/abs-1903-03096}, 
covers a wide range of 76 distinct task types within the field of natural language processing and comprises over 1,600 diverse NLP tasks.
During the experiments, 100 tasks were randomly selected, and for each task, 1000 samples were randomly selected for training and another 100 were selected for evaluation purpose. To ensure comparability, our sampling method follows the identical approach as described in~\citet{ponti2022combining}. To evaluate the effectiveness of the trained model, we employed various metrics for all selected tasks, including Exact Match (EM) and Rouge metrics~\citep{lin2004rouge}, including Rouge-1, Rouge-L, and Rouge-LSum.

To verify the universal effectiveness of our multitasking learning approach $\texttt{C-Poly}$, we chose T5 Version 1.1 - LM Adapted (T5)~\citep{raffel2020exploring}, FLAN-T5~\citep{chung2022scaling} and GLM~\citep{du2021glm} as the base models.

In our study, we thoroughly compared our proposed approach, $\texttt{C-Poly}$, and several existing MoE-like PEFT methods. The methods we compared against include LoRA, MoE-LoRA, $\texttt{Poly}$ and $\texttt{MHR}$. The comparison allowed us to analyze and evaluate the performance and effectiveness of our framework relative to these established PEFT methods.

\subsection{Training Details}
In our experiments, we applied PEFT methods to all query, key, value matrices within every attention layer in the base models. In the case of vanilla LoRA, we set the rank of the low-rank approximation, $r=8$. For all MoE-like tuning methods, we utilized in total 4 parallel LoRAs (experts) with $r=2$. In $\texttt{C-Poly}$, we set $A=3$ LoRA for task-common skills and $B=1$ LoRA for task-specific skills. This decision was made to ensure a comparable number of training parameters across all methods.

We trained our model for $1$ epoch only with a batch size of $4$ on both Super NI and SuperGLUE datasets during training. The AdamW optimizer~\citep{loshchilov2017decoupled} was used, with a learning rate of $5e^{-5}$. We also employed the linear decay strategy~\citep{loshchilov2016sgdr} as the learning rate scheduler with a weight decay of $0.01$ and a warmup ratio of $0.06$. All experiments were conducted on a single NVIDIA Tesla A100 graphics card.

\begin{figure}[htbp]
\begin{center}
\includegraphics[width=\textwidth]{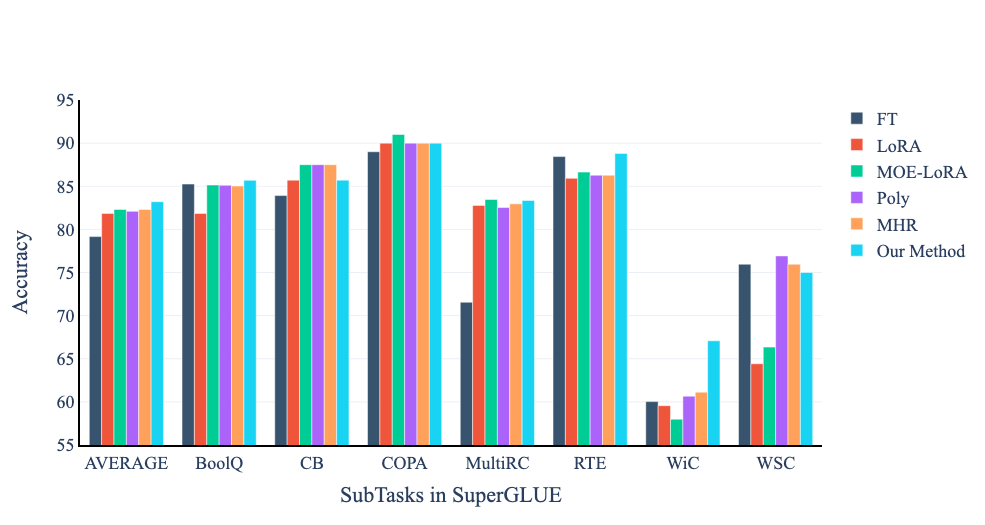}
\end{center}
\caption{FLAN-T5-Large with different PEFT methods on SuperGLUE benchmark, compared with Full Fine-tuning (FT). We reported overall averaged (AVERAGE) and task-specific accuracy for all sub-tasks.}
\label{fig:superglue-para}
\end{figure}

\subsection{Main Results and Discussion} 

\subsubsection{Analysis of Balanced Multitask Learning}

In Figure~\ref{fig:superglue-para}, we present a comparative analysis of various fine-tuning methods across multiple tasks within the SuperGLUE benchmark.
When tuning with full parameters (FT), the overall average accuracy is the lowest among all approaches because of relatively poor performance in the MultiRC sub-task.
This phenomenon, known as the seesaw effect, a manifestation of the negative transfer problem when tackling multiple tasks concurrently, holds significant importance in the domain of multi-task learning~\citep{2010A,2017Learning,10.1145/3383313.3412236}.
Our method $\texttt{C-Poly}$, on the other hand, demonstrates constant improvement over all sub-tasks thanks to the task-specific skill learning module.
The results reveal that $\texttt{C-Poly}$ can effectively mitigate the negative transfer and seesaw effect issues. As shown in Appendix~\ref{sec:A.4}, we conducted experiments on FLAN-T5-XL (2B)  and found that the base model with larger parameter values has a stronger fitting ability for multiple tasks. The experimental results showed that our method can bring stable improvements in reducing negative migration.
Consequently, our approach exhibits substantial superiority over other methods.

\begin{table}[htbp]
\caption{FLAN-T5-Large and GLM-10B with different adaptation methods on the SuperGLUE dataset. We report the overall (matched and mismatched) accuracy for BoolQ, CB, COPA, MultiRC, RTE, WiC and WSC. Higher is better for all metrics.}
\begin{center}
\label{tab:SuperGLUE}
\tabcolsep=0.11cm
\centering
\begin{tabular}{cccccccccc}
   \toprule
   \textbf{Base Model} & \textbf{PEFT Method} & \textbf{AVG} & \textbf{BoolQ} & \textbf{CB}& \textbf{COPA}& \textbf{MultiRC}& \textbf{RTE}& \textbf{WiC}& \textbf{WSC}\\ \midrule
   \multirow{5}{*}{FLAN-T5-Large}
   & LoRA & 81.85 & 81.85 & 85.71 & 90.00 & 82.78 & 85.92 & 59.56 & 64.42\\ 
   \cmidrule(lr){2-10}
   & MOE-LoRA & 82.31 & 85.14 & \textbf{87.50} & \textbf{91.00} & \textbf{83.46} & 86.64 & 57.99 & 66.35\\ 
   \cmidrule(lr){2-10}
   & $\texttt{Poly}$ & 82.09 & 85.11 & \textbf{87.50} & 90.00 & 82.53 & 86.28 & 60.66 & \textbf{76.92}\\
   \cmidrule(lr){2-10}
   & $\texttt{MHR}$ & 82.31 & 85.02 & \textbf{87.50} & 90.00 & 82.96 & 86.28 & 61.13 & 75.96\\ 
   \cmidrule(lr){2-10}
   & \textbf{Our Method} &\textbf{83.21} & \textbf{85.69} & 85.71 & 90.00 & 83.35 & \textbf{88.81} & \textbf{67.08} & 75.00\\
   \midrule
   \multirow{5}{*}{GLM-10B}
   & LoRA & 52.05 & 60.98 & 46.38 & 65.70 & 62.43 & 57.37 & 39.15 & 32.32\\ 
   \cmidrule(lr){2-10}
   & MoE-LoRA & 53.86 & 63.31 & 45.02 &	63.41 &	64.01 &	61.22 &	40.35 &	39.66\\ 
   \cmidrule(lr){2-10}
   & $\texttt{Poly}$ & 56.99 & 	64.65 &	52.17 &	65.54 &	65.66 &	62.15 &	41.71 &	47.08\\
   \cmidrule(lr){2-10}
   & $\texttt{MHR}$ & 56.92 & 64.85 & 50.79 & 66.36 & 65.79 & 62.75 & 42.35 & 45.58\\ 
   \cmidrule(lr){2-10}
   & \textbf{Our Method} &\textbf{62.26} & \textbf{67.31} & \textbf{60.38} &\textbf{70.04} & \textbf{67.90} & \textbf{68.01} & \textbf{48.71} & \textbf{53.42}\\
    \bottomrule
\end{tabular}
\end{center}
\end{table}

\subsubsection{Evaluation on More Models and More Tasks}
\label{sec:evaluation_on_more}

The effectiveness of different PEFT methods are evaluated on two architectures: T5 (Encoder-Decoder) and GLM (Decoder-Only). Table~\ref{tab:SuperGLUE} presents the performance of these methods on a dataset consisting of 7 tasks from the SuperGLUE. We display the evaluation results for the 7 sub-tasks individually, as well as their average performance. Table~\ref{tab:SuperNI} compares the performance of different PEFT methods on the SuperNI. The indicators in the table represent the average performance evaluation over 100 tasks.

The results are two-folded.
Firstly, it highlights that MoE-LoRA consistently demonstrates improvement over LoRA, attributed to the enhanced parameter flexibility from the MoE structure. We compared FLAN-T5-Large, T5-Large, and GLM-10B, and the results showed that it can significantly alleviate the phenomenon of negative migration in both encoder-decoder and decoder-only architectures and improve the flexibility of adaptive parameters compared to $\texttt{LoRA}$ and $\texttt{Poly}$. In the comparison of FLAN-T5-Large, although it has undergone a large number of pre-training with similar instruction samples, our method can still bring some improvement. The ablation experiments in Appendix~\ref{sec:A.2} on the T5 and FLAN-T5 series model of different scales (Large, XL, XXL) showed that our methods remained robust and significant as model parameters increased. 

Secondly, the results demonstrate that our method $\texttt{C-Poly}$ has achieved optimal performance in both architectures and on the SuperGLUE and SuperNI datasets. The ablation experiments in Appendix~\ref{sec:A.1} on the number of tasks (10-50-100) conducted on GLM-10B and FLAN-T5-Large indicate that our method is still significantly effective when the number of tasks increases. The explicit separation of task-specific skills and task-common skills in our design enables the skill modules to capture task-specific differences while sharing abstracted general skill modules effectively. Due to the explicit separation, the negative transfer phenomenon has been significantly reduced as in~\citet{10.1145/3383313.3412236}, which can be verified on both datasets and becomes more pronounced as the number of learning tasks increases. 

\begin{table}[]
\centering
\caption{FLAN-T5-Large, T5-Large and GLM-10B with different adaptation methods on the 100 randomly selected tasks from SuperNI dataset. We report the average Rouge1, RougeL, and RougeLsum for all tasks. Higher is better for all metrics.}
\label{tab:SuperNI}
\begin{tabular}{@{}lllll@{}}
\toprule
\textbf{Base Model} & \textbf{PEFT Method} & \textbf{rouge1} & \textbf{rougeL} & \textbf{rougeLsum} \\ \midrule
\multirow{5}{*}{FLAN-T5-Large} & LoRA                & 68.26          & 67.42          & 67.42          \\
                               & MoE-LoRA            & 68.59          & 67.76          & 67.75          \\
                               & Poly                & 68.45          & 67.60          & 67.58          \\
                               & MHR                 & 68.84          & 67.77          & 67.78          \\
                               & \textbf{Our Method} & \textbf{68.69} & \textbf{67.80} & \textbf{67.82} \\ \midrule
\multirow{5}{*}{T5-Large}      & LoRA                & 34.16          & 33.64          & 33.65          \\
                               & MoE-LoRA            & 36.82          & 36.13          & 36.15          \\
                               & Poly                & 43.04          & 42.05          & 42.09          \\
                               & MHR                 & 44.24          & 43.32          & 43.34          \\
                               & \textbf{Our Method} & \textbf{49.34} & \textbf{48.50} & \textbf{48.51} \\ \midrule
\multirow{5}{*}{GLM-10B}       & LoRA                & 43.16          & 42.04          & 42.09          \\
                               & MoE-LoRA            & 45.97          & 44.79          & 44.89          \\
                               & Poly                & 47.96          & 46.80          & 46.80          \\
                               & MHR                 & 48.53          & 47.34          & 47.33          \\
                               & \textbf{Our Method} & \textbf{49.53} & \textbf{48.45} & \textbf{48.45} \\ \bottomrule
\end{tabular}
\end{table}

\



\subsubsection{Parameter Efficiency Analysis}

\begin{figure}[H]
\begin{center}
\includegraphics[width=0.5\textwidth]{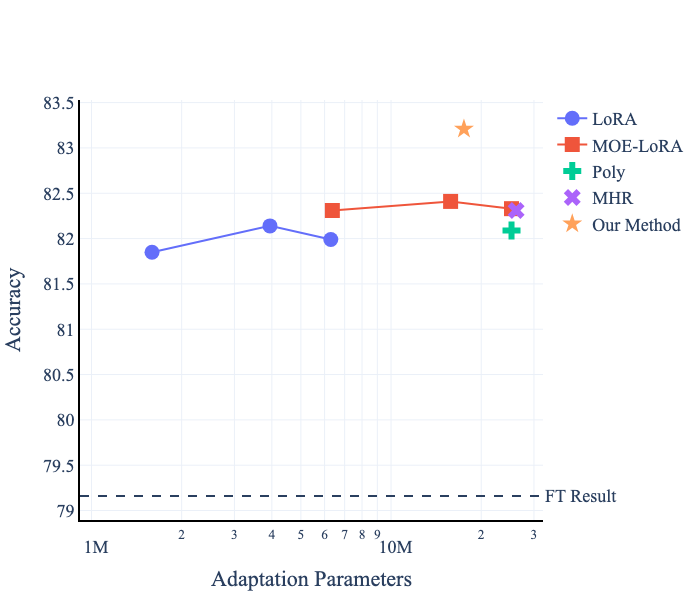}
\end{center}
\caption{Accuracy of PEFT methods and Full Fine-tuning on SuperGLUE dataset when applied on T5-Large. The X-axis shows the trainable parameter count during the fine-tuning process.}
\label{fig:superglue-acc}
\end{figure}

Figure~\ref{fig:superglue-acc} exhibits intriguing observations pertaining to various PEFT techniques within a constrained temporal epoch.
We assessed PEFT methodologies across varying parameter magnitudes, gauging their efficacy on the SuperGLUE multitask dataset with FLAN-T5-Large as the foundational model. 
Furthermore, we juxtaposed these outcomes against those derived from full parameter fine-tuning.
It is evident that our approach attains superior performance among all PEFT methodologies possessing identical parameter scales and training epochs.
Notably, $\texttt{C-Poly}$ even outperforms other methods with more parameters. Our methods explicitly segregate shared and proprietary skills and effectively ensure parameter efficiency in multi-task learning under equivalent training settings. 
This trend is also observed in the SuperNI benchmark evaluation.
As a result, we assert that adopting both task-common skills learning and task-specific skills learning as a paradigm constitutes a robust strategy for achieving parameter efficiency in adapting multiple tasks.
Furthermore, our approach surpasses the performance of full parameter fine-tuning while utilizing fewer parameters, exclusively updating routing parameters and adapters.

\subsubsection{Deeper Insights into $\texttt{C-Poly}$}

As outlined in Section~\ref{sec:evaluation_on_more}, our dual-skill framework, $\texttt{C-Poly}$, consistently delivers robust enhancements across various architectures, task scales, and model sizes. Here, we try to explore the intrinsic attributes of $\texttt{C-Poly}$ that drive these performance gains.

\textbf{Explicit skill separation enhances knowledge sharing }
The task-specific skills allow the task-common component to concentrate on the similarities across tasks, as the unique aspects of each task are handled by the task-specific skills.
Figures~\ref{fig:GLM_router_ni10}, ~\ref{fig:GLM_router_ni50}, and \ref{fig:GLM_router_ni100} demonstrate the task-common skill allocation $W_A$ in certain layers of the $\texttt{C-Poly}$ GLM-10B model for SuperNI-10, SuperNI-50, and SuperNI-100 respectively.
After proper normalization, these matrices reveal clear differences in skill allocation for distinct tasks. 
Additionally, we conducted task clustering based on the skill allocations learned in all layers of the GLM-10B model trained on SuperNI-100, with and without the task-specific component (i.e., comparing $\texttt{C-Poly}$ and $\texttt{Poly}$).
The clustering outcomes are shown as dendrograms in Figures\ref{fig:cpoly_GLM_100tasks} and ~\ref{fig:poly_GLM_100tasks}.
These results suggest that the enhanced performance of $\texttt{C-Poly}$ is due to a more balanced task hierarchy, facilitating more effective knowledge transfer among similar tasks and improved distinction of unrelated ones.

\textbf{Equilibrium in skill allocation is crucial }
The introduction of $\texttt{C-Poly}$ raises an important question about how to balance the allocation of parameters between task-common and task-specific skills. 
Specifically, we need to determine the optimal values of $A$ and $B$.
In Appendix~\ref{sec:A.3}, we conducted ablation experiments with different $(A, B)$ combinations while keeping the total trainable parameter count constant.
The findings corroborate our proposition that task-specific skills substantially elevate model efficacy. Nonetheless, it is equally vital to endow the task-common component with an adequate parameter count to successfully acquire shared knowledge.
Excessive focus on task-specific parameters can impede the learning process, potentially causing overfitting and hindering the model's ability to recognize similarities across tasks, which may lead to suboptimal performance.
A balanced parameter distribution promotes clear task distinction and helps prevent overfitting, preserving the model's generalization capabilities.
Identifying the ideal parameter ratio and configuration is a critical aspect for $\texttt{C-Poly}$ and may vary depending on the nature of the tasks and datasets involved.

\section{Related Works}
\subsection{Parameter-Efficient Fine-Tuning}
Numerous researchers have proposed incorporating adapters within neural networks, strategically placed between existing layers, and enforcing weight constraints on these adapters. LoRA~\citep{hu2022lora}, for instance, advocates for fine-tuning the model by learning low-rank matrix weights and aligning them with sovereignty. Building upon LoRA, (IA)$^3$~\citep{liu2022few} offers further enhancements by introducing a relatively modest number of novel parameters. In the context of model adjustment, prefix tuning emerges as a technique that exclusively optimizes a small segment of continuous task-specific vectors, thereby strengthening downstream task optimization.

\subsection{Modular Multi-task Learning}
Modular neural network architectures~\citep{jacobs1991task} have the advantages of positive transfer, compositionality, and parameter efficiency~\citep{zhang2023emergent}. Modular neural networks involve three identities: modules, routing function, and aggregation function. Modules represent different skills, which can be composed together and updated locally without affecting the rest of the network. A routing function determines the subset of modules for each task or example, which has variant types of fixed routing, learned hard routing, and learned soft routing~\citep{rosenbaum2017routing,jacobs1991adaptive,fernando2017pathnet}. An aggregating function determines a way to aggregate the outputs of active modules, which is implemented as a learnable neural network that depends on the output of all modules. Notably, within the realm of NLP, ~\citet{2101.03961}successfully extended the pre-training of large language models to trillions of parameters by leveraging the MoE architecture. Previous studies~\citep{1510.02879,ponti2022combining,2112.13208,kudugunta-etal-2021-beyond-distillation} have explored approaches to enforce parameter reuse and modularity in multitasking learning. ~\citet{1510.02879} trained individual modules for each task and subsequently learned how to reuse these modules. 

\subsection{Language Models}
The Transformer architecture serves as a fundamental framework for sequence pair modeling. Building upon this foundation, ~\citet{radford2018improving} employed a stack of Transformers to effectively model autoregressive languages through the deployment of encoders and decoders. BERT~\citep{devlin2019bert} and GPT-2~\citep{radford2019language} are classic text modeling methodologies, both relying on Transformer units pre-trained on vast amounts of textual data. The encoder-only architecture model is particularly suited for comprehension-based tasks, whereas generative tasks benefit from both encoder-decoder and decoder-only architecture models due to their autoregressive nature~\citep{fu2023decoderonly,sarrouti-etal-2022-comparing}.

\section{Conclusion}
\label{conclusion}

In this article, we introduce a novel paradigm for PEFT called Customized Polytropon $\texttt{C-Poly}$.
By explicitly distinguishing between task-common and task-specific skills, our method enables efficient multi-task fine-tuning on large language models, even with limited computational resources.
Our approach addresses the challenge of resource limitations and allows for efficient training. 
The separation of exclusive and general skills effectively mitigates the seesaw problem and negative transfer commonly encountered in multitasking learning, leading to superior overall performance, which also offers compelling interpretability.
Extensive experimental evaluations demonstrate the effectiveness of our proposed method, surpassing existing PEFT baselines and achieving state-of-the-art performance.
These results highlight the potential and significance of our unified multi-task learning framework $\texttt{C-Poly}$ in the field of parameter-efficient multi-task learning.

\newpage
\bibliography{iclr2024_conference}
\bibliographystyle{iclr2024_conference}

\appendix

\section{Additional results}
\subsection{PEFT Performance on different task numbers}
\label{sec:A.1}
We compared and evaluated the performance of different PEFT multitasking tuning frameworks on different bases and with different task quantities. Table~\ref{tab:t5_varies_tasks_experiment}, Table~\ref{tab:flan_t5_varies_tasks_experiment} and Table~\ref{tab:glm_10B_varies_tasks_experiment} show the performance of multi task tuning with different task quantities based on T5-Large, Flan-T5-Large, and GLM-10B models respectively. It can be clearly seen that our method can achieve significant improvements under different numbers of tasks.

It is worth mentioning that FLAN-T5-Large is a model obtained from large-scale training of the T5-Large architecture model on FLAN data, and its training set may have some similarities or overlaps with the NI multitasking instruction data training set, which makes its performance very high and reduces the gap between different PEFT methods.

\begin{table}[htbp]
\centering
\caption{Performance on T5-Large with different numbers of randomly
selected tasks from SuperNI dataset. We report the average Rouge1, RougeL, and RougeLsum for
all tasks. Higher is better for all metrics}
\label{tab:t5_varies_tasks_experiment}
\begin{tabular}{@{}lllll@{}}
\toprule
\textbf{Tasks Numbers}                      & \textbf{PEFT Method}        & \textbf{rouge1}              & \textbf{rougeL}              & \textbf{rougeLsum}           \\ \midrule
{\color[HTML]{000000} }                     & {\color[HTML]{000000} LoRA} & {\color[HTML]{000000} 14.22} & {\color[HTML]{000000} 14.12} & {\color[HTML]{000000} 14.19} \\
{\color[HTML]{000000} }                     & MoE-LoRA                    & 16.75                        & 16.71                        & 16.76                        \\
{\color[HTML]{000000} }                     & {\color[HTML]{000000} Poly} & {\color[HTML]{000000} 17.34} & {\color[HTML]{000000} 17.31} & {\color[HTML]{000000} 17.38} \\
{\color[HTML]{000000} }                     & {\color[HTML]{000000} MHR}  & {\color[HTML]{000000} 17.17} & {\color[HTML]{000000} 17.11} & {\color[HTML]{000000} 17.16} \\
\multirow{-5}{*}{{\color[HTML]{000000} 10}} & \textbf{Our Method}         & \textbf{42.62}               & \textbf{42.47}               & \textbf{42.60}               \\ \midrule
                                            & LoRA                        & 32.58                        & 31.64                        & 31.59                        \\
                                            & MoE-LoRA                    & 39.50                        & 38.47                        & 38.46                        \\
                                            & Poly                        & 46.13                        & 44.25                        & 44.28                        \\
                                            & MHR                         & 47.27                        & 45.43                        & 45.39                        \\
\multirow{-5}{*}{50}                        & \textbf{Our Method}         & \textbf{53.39}               & \textbf{51.68}               & \textbf{51.63}               \\ \midrule
                                            & LoRA                        & 34.16                        & 33.64                        & 33.65                        \\
                                            & MoE-LoRA                    & 36.82                        & 36.13                        & 36.15                        \\
                                            & Poly                        & 43.04                        & 42.05                        & 42.09                        \\
                                            & MHR                         & 44.24                        & 43.32                        & 43.34                        \\
\multirow{-5}{*}{100}                       & \textbf{Our Method}         & \textbf{49.34}               & \textbf{48.50}               & \textbf{48.51}               \\ \bottomrule
\end{tabular}
\end{table}
\begin{table}[htbp]
\centering
\caption{Performance on FLAN-T5-Large with different numbers of randomly
selected tasks from SuperNI dataset. We report the average Rouge1, RougeL, and RougeLsum for
all tasks. Higher is better for all metrics}
\label{tab:flan_t5_varies_tasks_experiment}
\begin{tabular}{@{}lllll@{}}
\toprule
\textbf{Tasks Numbers}                      & \textbf{PEFT Method}        & \textbf{rouge1}              & \textbf{rougeL}              & \textbf{rougeLsum}           \\ \midrule
{\color[HTML]{000000} }                     & {\color[HTML]{000000} LoRA} & {\color[HTML]{000000} 67.82} & {\color[HTML]{000000} 67.01} & {\color[HTML]{000000} 67.03} \\
{\color[HTML]{000000} }                     & MoE-LoRA                    & 67.95                        & 67.12                        & 67.15                        \\
{\color[HTML]{000000} }                     & {\color[HTML]{000000} Poly} & {\color[HTML]{000000} 68.10} & {\color[HTML]{000000} 67.29} & {\color[HTML]{000000} 67.33} \\
{\color[HTML]{000000} }                     & {\color[HTML]{000000} MHR}  & {\color[HTML]{000000} 77.49} & {\color[HTML]{000000} 77.25} & {\color[HTML]{000000} 77.36} \\
\multirow{-5}{*}{{\color[HTML]{000000} 10}} & \textbf{Our Method}         & \textbf{77.73}               & \textbf{77.58}               & \textbf{77.61}               \\ \midrule
                                            & LoRA                        & 70.66                        & 69.10                        & 69.03                        \\
                                            & MoE-LoRA                    & 70.81                        & 69.25                        & 69.21                        \\
                                            & Poly                        & 70.76                        & 69.23                        & 69.15                        \\
                                            & MHR                         & 70.92                        & 69.39                        & 69.33                        \\
\multirow{-5}{*}{50}                        & \textbf{Our Method}         & \textbf{71.17}               & \textbf{69.68}               & \textbf{69.62}               \\ \midrule
                                            & LoRA                        & 68.26                        & 67.42                        & 67.42                        \\
                                            & MoE-LoRA                    & 68.59                        & 67.76                        & 67.75                        \\
                                            & Poly                        & 68.45                        & 67.60                        & 67.58                        \\
                                            & MHR                         & 68.84                        & 67.77                        & 67.78                        \\
\multirow{-5}{*}{100}                       & \textbf{Our Method}         & \textbf{68.69}               & \textbf{67.80}               & \textbf{67.82}               \\ \bottomrule
\end{tabular}
\end{table}

\begin{table}[htbp]
\centering
\caption{Performance on GLM-10B with different numbers of randomly
selected tasks from SuperNI dataset. We report the average Rouge1, RougeL, and RougeLsum for
all tasks. Higher is better for all metrics}
\label{tab:glm_10B_varies_tasks_experiment}
\begin{tabular}{@{}lllll@{}}
\toprule
\textbf{Tasks Numbers}                      & \textbf{PEFT Method}        & \textbf{rouge1}              & \textbf{rougeL}              & \textbf{rougeLsum}           \\ \midrule
{\color[HTML]{000000} }                     & {\color[HTML]{000000} LoRA} & {\color[HTML]{000000} 30.64} & {\color[HTML]{000000} 30.40} & {\color[HTML]{000000} 30.42} \\
{\color[HTML]{000000} }                     & MoE-LoRA                    & 33.92                        & 33.79                        & 33.77                        \\
{\color[HTML]{000000} }                     & {\color[HTML]{000000} Poly} & {\color[HTML]{000000} 34.53} & {\color[HTML]{000000} 34.41} & {\color[HTML]{000000} 34.31} \\
{\color[HTML]{000000} }                     & {\color[HTML]{000000} MHR}  & {\color[HTML]{000000} 33.63} & {\color[HTML]{000000} 33.47} & {\color[HTML]{000000} 33.47} \\
\multirow{-5}{*}{{\color[HTML]{000000} 10}} & \textbf{Our Method}         & \textbf{43.74}               & \textbf{43.72}               & \textbf{43.65}               \\ \midrule
                                            & LoRA                        & 34.16                        & 33.00                        & 32.98                        \\
                                            & MoE-LoRA                    & 39.87                        & 38.63                        & 38.55                        \\
                                            & Poly                        & 44.81                        & 43.09                        & 43.07                        \\
                                            & MHR                         & 45.32                        & 43.62                        & 43.56                        \\
\multirow{-5}{*}{50}                        & \textbf{Our Method}         & \textbf{53.17}               & \textbf{51.27}               & \textbf{51.32}               \\ \midrule
                                            & LoRA                        & 43.16                        & 42.04                        & 42.09                        \\
                                            & MoE-LoRA                    & 45.97                        & 44.79                        & 44.89                        \\
                                            & Poly                        & 47.96                        & 46.80                        & 46.80                        \\
                                            & MHR                         & 48.53                        & 47.34                        & 47.33                        \\
\multirow{-5}{*}{100}                       & \textbf{Our Method}         & \textbf{49.53}               & \textbf{48.45}               & \textbf{48.45}               \\ \bottomrule
\end{tabular}
\end{table}

\subsection{PEFT performances on different scale of base model}
\label{sec:A.2}
Table~\ref{tab:T5_Architecture_100tasks_experiment} and Table~\ref{tab:FLAN_T5_Architecture_100tasks_experiment} show the performance of T5 and FLAN-T5 models with different parameter quantities under different PEFT methods, respectively. It can be seen that our method has a robust improvement on models with different parameter quantities, and as the parameter quantity increases, the magnitude of the improvement gradually increases.

\begin{table}[H]
\centering
\caption{T5 with different parameter scales' performances with different adaptation methods on the 100 randomly
selected tasks from SuperNI dataset. We report the average Rouge1, RougeL, and RougeLsum for
all tasks. Higher is better for all metrics}
\label{tab:T5_Architecture_100tasks_experiment}
\begin{tabular}{@{}cllll@{}}
\toprule
\multicolumn{1}{l}{Base Model} & \textbf{PEFT Method} & \textbf{Rouge1} & \textbf{RougeL} & \textbf{RougeLsum} \\ \midrule
\multirow{5}{*}{T5-Large} & LoRA                & 34.16          & 33.64          & 33.65          \\
                          & MoE-LoRA            & 36.82          & 36.13          & 36.15          \\
                          & Poly                & 43.04          & 42.05          & 42.09          \\
                          & MHR                 & 44.24          & 43.32          & 43.34          \\
                          & \textbf{Our Method} & \textbf{49.34} & \textbf{48.50} & \textbf{48.51} \\ \midrule
\multirow{5}{*}{T5-XL}    & LoRA                & 34.93          & 34.34          & 34.40          \\
                          & MoE-LoRA            & 39.78          & 38.83          & 38.87          \\
                          & Poly                & 43.61          & 42.61          & 42.62          \\
                          & MHR                 & 45.53          & 44.62          & 44.61          \\
                          & \textbf{Our Method} & \textbf{50.57} & \textbf{49.74} & \textbf{49.76} \\ \midrule
\multirow{5}{*}{T5-XXL}   & LoRA                & 49.97          & 48.89          & 48.93          \\
                          & MoE-LoRA            & 52.14          & 51.12          & 51.15          \\
                          & Poly                & 55.42          & 54.65          & 54.64          \\
                          & MHR                 & 55.81          & 55.01          & 55.01          \\
                          & \textbf{Our Method} & \textbf{62.23} & \textbf{61.44} & \textbf{61.47} \\ \bottomrule
\end{tabular}
\end{table}

\begin{table}[H]
\centering
\caption{FLAN-T5 with different parameter scales' performances with different adaptation methods on the 100 randomly
selected tasks from SuperNI dataset. We report the average Rouge1, RougeL, and RougeLsum for
all tasks. Higher is better for all metrics}
\label{tab:FLAN_T5_Architecture_100tasks_experiment}
\begin{tabular}{@{}cllll@{}}
\toprule
\multicolumn{1}{l}{Base Model} & \textbf{PEFT Method} & \textbf{Rouge1} & \textbf{RougeL} & \textbf{RougeLsum} \\ \midrule
\multirow{5}{*}{FLAN-T5-Large} & LoRA                & 68.26          & 67.42          & 67.42          \\
                               & MoE-LoRA            & 68.59          & 67.76          & 67.75          \\
                               & Poly                & 68.45          & 67.60          & 67.58          \\
                               & MHR                 & 68.84          & 67.77          & 67.78          \\
                               & \textbf{Our Method} & \textbf{68.69} & \textbf{67.80} & \textbf{67.82} \\ \midrule
\multirow{5}{*}{FLAN-T5-XL}    & LoRA                & 71.01          & 70.21          & 70.24          \\
                               & MoE-LoRA            & 71.08          & 70.29          & 70.33          \\
                               & Poly                & 71.12          & 70.31          & 70.35          \\
                               & MHR                 & 71.18          & 70.36          & 70.40          \\
                               & \textbf{Our Method} & \textbf{71.57} & \textbf{70.72} & \textbf{70.74} \\ \midrule
\multirow{5}{*}{FLAN-T5-XXL}   & LoRA                & 71.89          & 71.07          & 71.08          \\
                               & MoE-LoRA            & 72.05          & 71.25          & 71.26          \\
                               & Poly                & 72.55          & 71.76          & 71.78          \\
                               & MHR                 & 72.40          & 71.61          & 71.63          \\
                               & \textbf{Our Method} & \textbf{73.09} & \textbf{72.27} & \textbf{72.28} \\ \bottomrule
\end{tabular}
\end{table}

\subsection{PEFT Experiment with different numbers of Common/task-specific skills}
\label{sec:A.3}
We designed an experiment to investigate the impact of the relative size of common skill and task specific skill on the results under the same parameter quantity. From the table~\ref{tab:common_specific_100tasks_experiment}, it can be seen that for different base models, the gain is most significant when the number of task specific skills is 1, indicating that explicitly separating the general skills and proprietary skills of skills is the main factor that brings improvement, which is consistent with the methods and viewpoints explained in our paper.

\begin{table}[H]
\centering
\caption{Common/task-specific skills experiment on T5-Large, FLAN-T5-Large and GLM-10B with different adaptation methods on the 100 randomly
selected tasks from SuperNI dataset. We report the average Rouge1, RougeL, and RougeLsum for
all tasks. Higher is better for all metrics}
\label{tab:common_specific_100tasks_experiment}
\begin{tabular}{@{}llllll@{}}
\toprule
\textbf{Base Model} &
  \textbf{N Common Skills} &
  \textbf{N Specific Skills} &
  \textbf{Rouge1} &
  \textbf{RougeL} &
  \textbf{RougeLsum} \\ \midrule
{\color[HTML]{000000} } &
  {\color[HTML]{000000} 4} &
  {\color[HTML]{000000} 0} &
  {\color[HTML]{000000} 43.04} &
  {\color[HTML]{000000} 42.05} &
  {\color[HTML]{000000} 42.09} \\
{\color[HTML]{000000} } &
  {\color[HTML]{000000} 3} &
  {\color[HTML]{000000} 1} &
  {\color[HTML]{000000} 49.34} &
  {\color[HTML]{000000} 48.50} &
  {\color[HTML]{000000} 48.51} \\
{\color[HTML]{000000} } &
  {\color[HTML]{000000} 2} &
  {\color[HTML]{000000} 2} &
  {\color[HTML]{000000} 48.79} &
  {\color[HTML]{000000} 47.94} &
  {\color[HTML]{000000} 47.94} \\
\multirow{-4}{*}{{\color[HTML]{000000} T5-Large}} &
  {\color[HTML]{000000} 1} &
  {\color[HTML]{000000} 3} &
  {\color[HTML]{000000} 47.46} &
  {\color[HTML]{000000} 46.60} &
  {\color[HTML]{000000} 46.59} \\ \midrule
 &
  4 &
  0 &
  68.45 &
  67.60 &
  67.58 \\
 &
  3 &
  1 &
  68.69 &
  67.80 &
  67.82 \\
 &
  2 &
  2 &
  68.47 &
  67.63 &
  67.65 \\
\multirow{-4}{*}{FLAN-T5-Large} &
  1 &
  3 &
  68.21 &
  67.36 &
  67.41 \\ \midrule
 &
  4 &
  0 &
  47.96 &
  46.80 &
  46.80 \\
 &
  3 &
  1 &
  49.53 &
  48.45 &
  48.45 \\
 &
  2 &
  2 &
  49.19 &
  48.16 &
  48.16 \\
\multirow{-4}{*}{GLM-10B} &
  1 &
  3 &
  46.85 &
  45.74 &
  45.73 \\ \bottomrule
\end{tabular}
\end{table}

\subsection{Model Scale Experiment on SuperGLUE}
\label{sec:A.4}
\begin{figure}[htbp]
\begin{center}
\includegraphics[width=\textwidth]{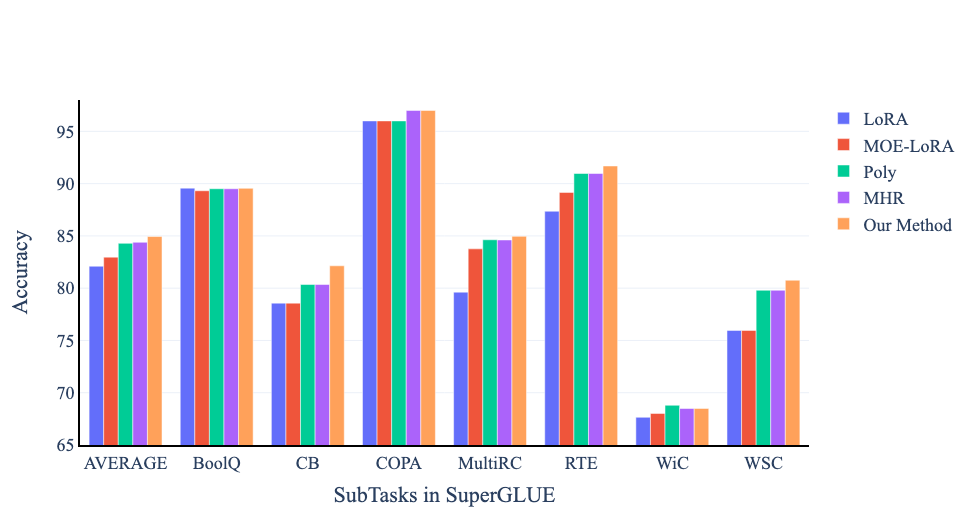}
\end{center}
\caption{FLAN-T5-XL with different PEFT methods on SuperGLUE benchmark. We reported overall averaged (AVERAGE) and task-specific accuracy for all sub-tasks.}
\label{fig:superglue-flant5-xl}
\end{figure}

We conduct SuperGLUE experiment on FLAN-T5-XL as shown in Figure~\ref{fig:superglue-flant5-xl} and Table~\ref{tab:flan_t5_xl_superglue} to show the performance of different PEFT methods. Experiments have shown that compared to FLAN-T5-Large (0.78B) in Figure~\ref{fig:superglue-para}, FLAN-T5-XL (2B) can achieve a more stable improvement on the SuperGLUE dataset after increasing the parameter quantity, this indicates that as the number of parameters increases, the model's fitting ability becomes stronger and can adapt to the parameter learning needs of multiple tasks, while our method weakens the occurrence of negative transfer phenomenon.

\begin{table}[H]
\centering
\caption{FLAN-T5-XL with different adaptation methods on the SuperGLUE
dataset. We report the overall (matched and mismatched) accuracy for BoolQ, CB, COPA, MultiRC,
RTE, WiC and WSC. Higher is better for all metrics.}
\label{tab:flan_t5_xl_superglue}
\begin{tabular}{@{}lllllllll@{}}
\toprule
                    & \textbf{avg}   & \textbf{boolq} & \textbf{cb}    & \textbf{copa}  & \textbf{multirc} & \textbf{rte}   & \textbf{wic} & \textbf{wsc.fixed} \\ \midrule
lora      & 82.10 & \textbf{89.57} & 78.57 & 96.00 & 79.62 & 87.36 & 67.65          & 75.96 \\
moe-lora  & 82.97 & 89.32          & 78.57 & 96.00 & 83.76 & 89.16 & 68.02          & 75.96 \\
poly-lora & 84.29 & 89.51          & 80.35 & 96.00 & 84.63 & 90.97 & \textbf{68.80} & 79.80 \\
mhr-lora  & 84.39 & 89.51          & 80.35 & 97.00 & 84.61 & 90.97 & 68.49          & 79.80 \\
\textbf{cpoly-lora} & \textbf{84.94} & 89.55          & \textbf{82.14} & \textbf{97.00} & \textbf{84.96}   & \textbf{91.69} & 68.50        & \textbf{80.76}     \\ \bottomrule
\end{tabular}
\end{table}

\subsection{Learned Common-Skill Router Visualization and Analysis}
\label{sec:A.5}
\begin{figure}[htbp]
\begin{center}
\includegraphics[width=\textwidth]{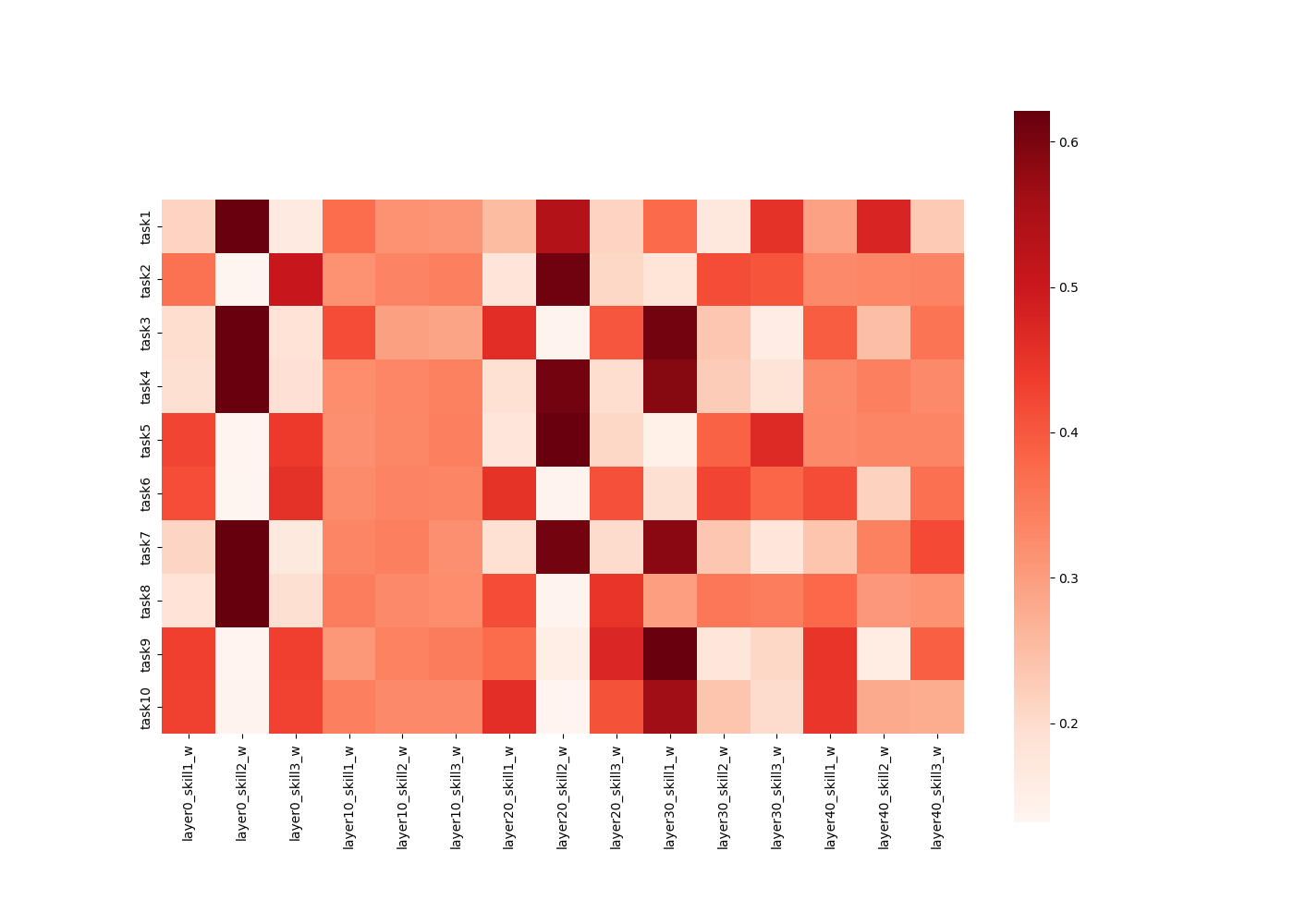}
\end{center}
\caption{Visualization of the common skill allocation matrix $W_A$ of selected transformer layers in GLM-10B after training on SuperNI-10.}
\label{fig:GLM_router_ni10}
\end{figure}

\begin{figure}[htbp]
\begin{center}
\includegraphics[width=0.7\textwidth]{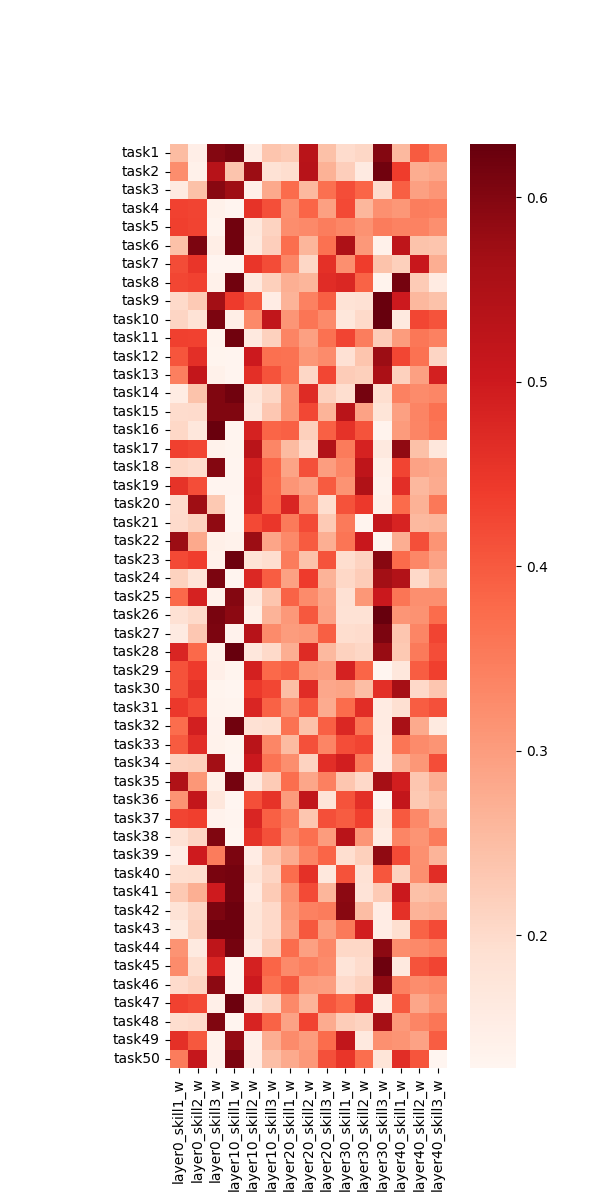}
\end{center}
\caption{Visualization of the common skill allocation matrix $W_A$ of selected transformer layers in GLM-10B after training on SuperNI-50.}
\label{fig:GLM_router_ni50}
\end{figure}

\begin{figure}[htbp]
\begin{center}
\includegraphics[width=0.7\textwidth]{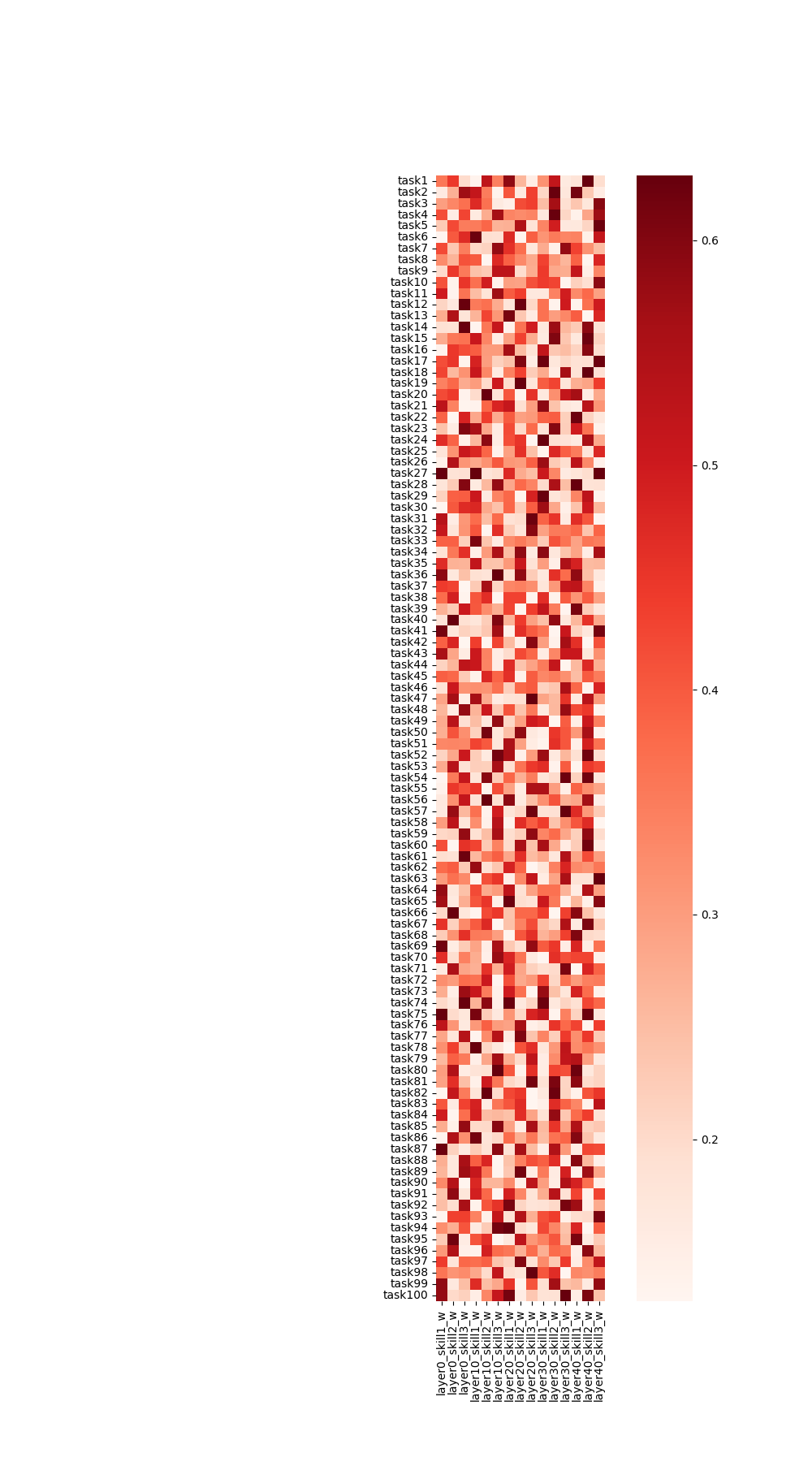}
\end{center}
\caption{Visualization of the common skill allocation matrix $W_A$ of selected transformer layers in GLM-10B after training on SuperNI-100.}
\label{fig:GLM_router_ni100}
\end{figure}

Figure~\ref{fig:GLM_router_ni10}, Figure~\ref{fig:GLM_router_ni50}, and Figure~\ref{fig:GLM_router_ni100} shows the distribution of router weights of common skill for selected transformers layers (we select layer0-10-20-30-40 of transformer layers in GLM-10B Model)
in NI-10, NI-50, NI-100 Tasks experiment of GLM-10B, corresponding to the three sets of experiments in Appendix: Table~\ref{tab:glm_10B_varies_tasks_experiment}. The allocation of weights shows that the learned router vector weights for different tasks are clearly separated and differentiated.

\begin{figure}[H]
\begin{center}
\includegraphics[width=\textwidth]{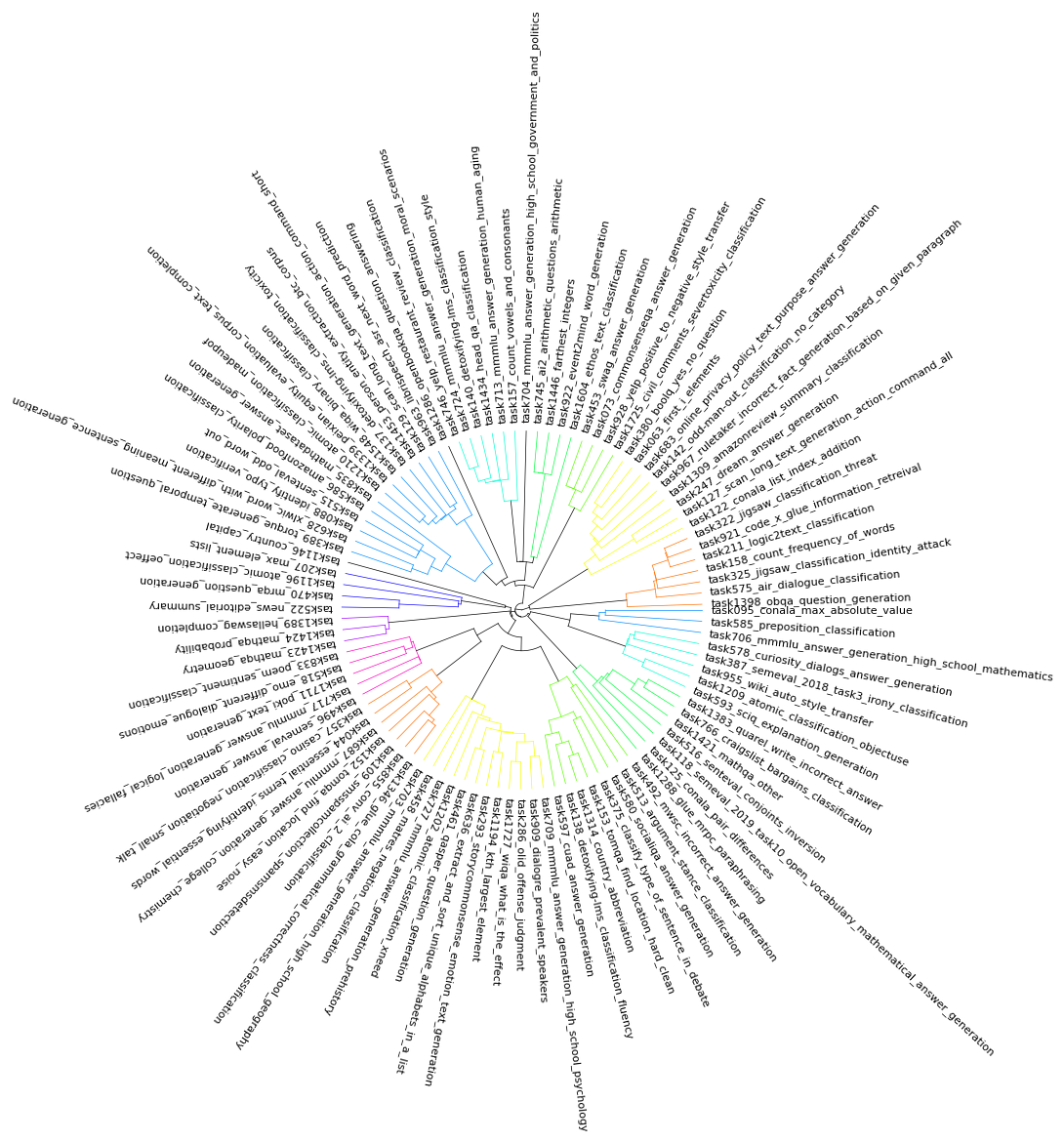}
\end{center}
\caption{
Task clustering dendrogram for common skill allocation matrix $W_A$ of $\texttt{C-Poly}$ using GLM-10B as the base model in the NI-100-Tasks experiment.
Tasks are grouped into the same category if they share a similar subset of skills.
}
\label{fig:cpoly_GLM_100tasks}
\end{figure}

\begin{figure}[H]
\begin{center}
\includegraphics[width=\textwidth]{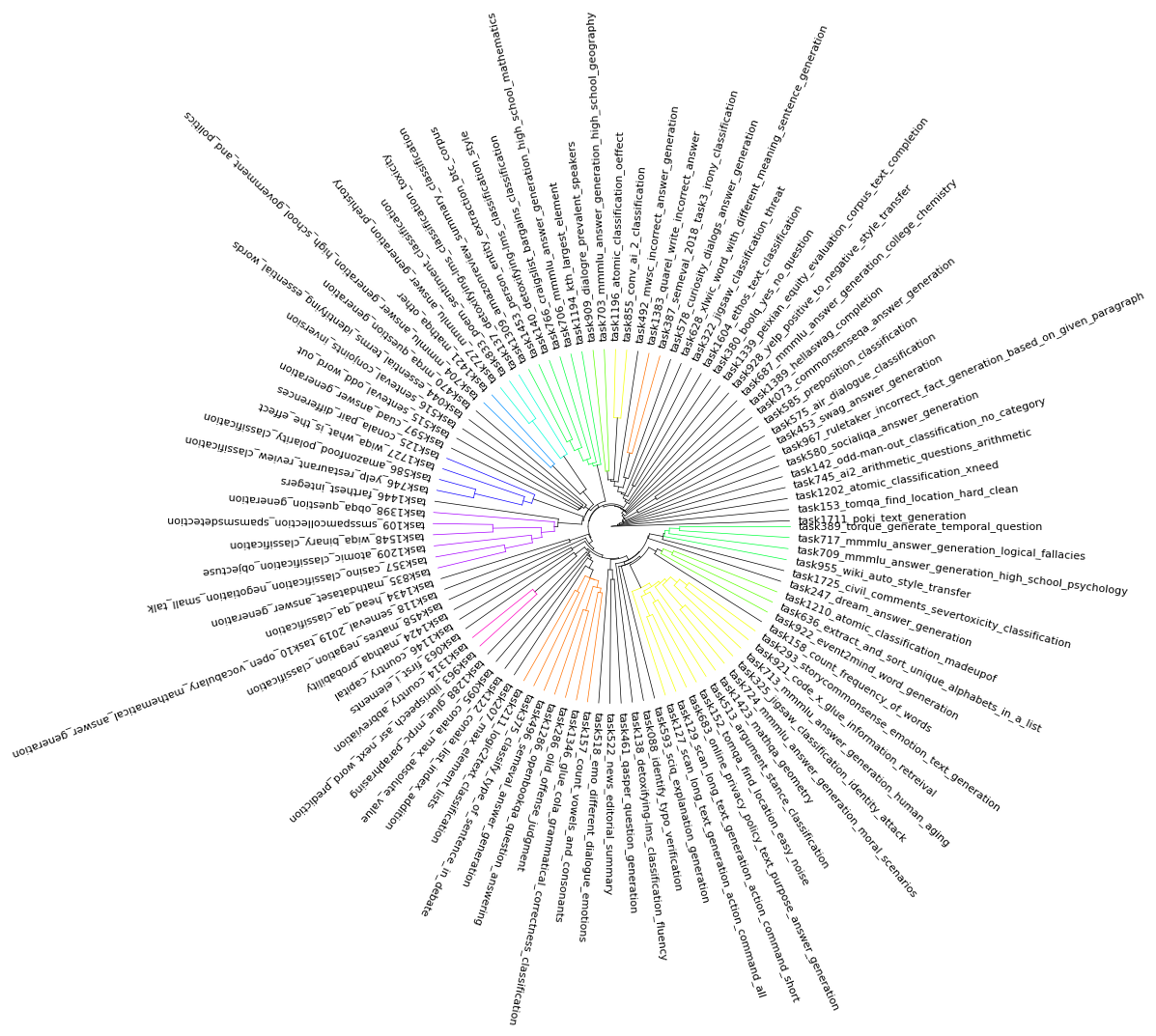}
\end{center}
\caption{Task clustering dendrogram for common skill allocation matrix $W_A$ of $\texttt{Poly}$ using GLM-10B as the base model in NI-100-Tasks experiment.
Tasks are grouped into the same category if they share a similar subset of skills.}
\label{fig:poly_GLM_100tasks}
\end{figure}

Figure~\ref{fig:cpoly_GLM_100tasks} and Figure~\ref{fig:poly_GLM_100tasks} show that our method achieves stronger learning ability in the common-skill router after independently splitting task-specific skills for individual learning, which allows for more discriminative learning of differences and similarities among tasks. This is reflected in the more reasonable types of tasks clustered by our router, with a more balanced number of tasks in each abstract class and more sufficient learning than the origin poly.

\end{document}